\documentclass[twoside,11pt]{article}

\usepackage{blindtext}

\usepackage[abbrvbib,preprint]{dmlr2e}

\usepackage{hyperref}
\usepackage{acronym}
\usepackage{amsmath}
\usepackage{amssymb}
\usepackage{xcolor}
\usepackage{booktabs}
\usepackage{placeins}

\graphicspath{{images/}}

\usepackage{caption}
\usepackage{subcaption}

\newcommand{\pic}[3]{
	\begin{figure}[tb] 
		\centering
		\includegraphics[width=#3\linewidth]{#1}
		\caption{#2}
		\label{fig:#1}
	\end{figure}
}

\newcommand{\picLarge}[3]{
	\begin{figure*}[tb] 
		\centering
		\includegraphics[width=#3\linewidth]{#1}
		\caption{#2}
		\label{fig:#1}
	\end{figure*}
}

\newcommand{\picTwoVertical}[4]{
	\begin{figure*}[tb]
		\centering
		\begin{subfigure}[c]{0.45\textwidth}
			\centering		
			\includegraphics[width=0.9\textwidth]{#1-0}
			\subcaption{#2}	
			\label{fig:#1-0}	
		\end{subfigure}
		\begin{subfigure}[c]{0.45\textwidth}	
			\centering	
			\includegraphics[width=0.9\textwidth]{#1-1}
			\subcaption{#3}
			\label{fig:#1-1}	
		\end{subfigure}
		\caption{#4}
		\label{fig:#1}
	\end{figure*}
}

\newcommand{\picFour}[6]{
	\begin{figure}[tb]
		\centering
		\begin{subfigure}[c]{0.24\textwidth}
			\centering		
			\includegraphics[width=0.9\textwidth]{#1-0}
			\subcaption{#2}	
			\label{fig:#1-0}	
		\end{subfigure}
		\begin{subfigure}[c]{0.24\textwidth}	
			\centering	
			\includegraphics[width=0.9\textwidth]{#1-1}
			\subcaption{#3}
			\label{fig:#1-1}	
		\end{subfigure}
  \begin{subfigure}[c]{0.24\textwidth}
			\centering		
			\includegraphics[width=0.9\textwidth]{#1-2}
			\subcaption{#4}	
			\label{fig:#1-2}	
		\end{subfigure}
		\begin{subfigure}[c]{0.24\textwidth}	
			\centering	
			\includegraphics[width=0.9\textwidth]{#1-3}
			\subcaption{#5}
			\label{fig:#1-3}	
		\end{subfigure}
		\caption{#6}
		\label{fig:#1}
	\end{figure}
}

\newcommand{\tblDeepLearningResults}{    
\begin{table*}
        \centering
        \caption{Results of KL divergence across different methods and number of annotations -- 
        The first method follows our recommended guidelines.
        The next two methods utilize proposals, while the last three do not, leading to a slower annotation time compared to the recommendation.
        Therefore, when comparing, one should consider a higher number of annotations for DC3 proposals, as they could potentially be achieved in the same or less time.
        The results are presented as KL $\pm$ STD (Relative Change in \% compared to the recommended method).}
        \label{tbl:dlResults} 
        
        \resizebox{\linewidth}{!}{%
        \begin{tabular}{l c  c    c c c c}
            \toprule
            \multicolumn{3}{c}{Method} & \multicolumn{4}{c}{Number of Annotations} \\ 
            \cmidrule(r){1-3} \cmidrule(r){4-7}
            Proposal             &   Blending & Correction & 1 & 3 & 5 & 10 \\
            \midrule
            DC3      &    Balanced & Yes &0.4481 $\pm$ 0.1957 (0.00\%) & 0.2425 $\pm$ 0.0741 (0.00\%) & 0.1832 $\pm$ 0.0095 (0.00\%) & 0.1615 $\pm$ 0.0129 (0.00\%) \\
            \midrule
            DC3       &    Balanced & No  &1.0309 $\pm$ 0.1750 (130.07\%) & 0.2704 $\pm$ 0.0760 (11.53\%) & 0.2105 $\pm$ 0.0256 (14.85\%) & 0.4546 $\pm$ 0.4343 (181.49\%)  \\
            DC3      &   No        & Yes &1.7382 $\pm$ 0.1553 (287.91\%) & 0.2499 $\pm$ 0.0374 (3.08\%) & 0.2081 $\pm$ 0.0171 (13.57\%) & 0.2798 $\pm$ 0.1325 (73.25\%) \\
            \midrule
            No                   & Only Blends & No  &0.2311 $\pm$ 0.0224 (-48.42\%) & 0.2394 $\pm$ 0.0930 (-1.27\%) & 0.1904 $\pm$ 0.0209 (3.88\%) & 0.1644 $\pm$ 0.0062 (1.79\%) \\
            No                   &    Balanced & No  &0.8565 $\pm$ 0.4466 (91.14\%) & 0.1966 $\pm$ 0.0245 (-18.90\%) & 0.1678 $\pm$ 0.0224 (-8.44\%) & 0.1568 $\pm$ 0.0165 (-2.92\%)  \\
            No                   &   No        & No  &0.5578 $\pm$ 0.1259 (24.49\%) & 0.2451 $\pm$ 0.0127 (1.10\%) & 0.5435 $\pm$ 0.4285 (196.61\%) & 0.1898 $\pm$ 0.0380 (17.52\%)  \\
            \bottomrule
        \end{tabular}
        }
\end{table*}
}

\newcommand{\tblhumanResults}{    
\begin{table}
        \centering
        \caption{Comparison of human and network performance on Verse2019~\citep{Loffler2020Verse2019} -- * based on ten randomly sampled annotations}
        \label{tbl:originalResults} 
        
        \resizebox{0.53\linewidth}{!}{%
        \subfloat[Human Performance]{
            \begin{tabular}{l c}
                \toprule
                &  $F1$ \\
                \midrule
                2D - All & 0.6380 $\pm$ 0.0323 \\ 
                2D - No Proposal & 0.6297  $\pm$ 0.0345 \\
                2D - With Proposal & 0.6462 $\pm$ 0.0302 \\ 
                \midrule 
                3D - No Proposal & 0.6243 $\pm$ 0.0370 \\
                \bottomrule
            \end{tabular}
        }
    }
    \quad
    \resizebox{0.42\linewidth}{!}{%
        \subfloat[Network]{
            \begin{tabular}{l c}
                \toprule
                &  $F1$ \\
                \midrule
                Sampled from Human* & 0.69 \\
                \midrule
                Proposal Network & 0.57 \\ 
                Wei et al.~\citep{Wei2022Verse} & 0.59 \\ 
                Ours* & 0.63 \\
                \bottomrule
            \end{tabular}
        }
    }
\end{table}
}

\usepackage{lastpage}
\dmlrheading{24}{2024}{1-\pageref{LastPage}}{01/24; Revised 03/24}{05/24}{21-0000}{Lars Schmarje, Vasco Grossmann, Claudius Zelenka, Johannes Brünger, Reinhard Koch
} 
\def\openreview{\url{https://openreview.net/forum?id=gii9qgtvsL}} 

\ShortHeadings{Annotating Ambiguous Images}{General Annotation Strategy for High-Quality Data}
\firstpageno{1}

\begin{document}

\title{Annotating Ambiguous Images:  General Annotation Strategy for High-Quality Data with Real-World Biomedical Validation}

\author{\name Lars Schmarje
\email science@schmarje-sh.de
 \addr Kiel University
\AND Vasco Grossmann
\email vgr@informatik.uni-kiel.de
\addr Kiel University
\AND Claudius Zelenka
\email cze@informatik.uni-kiel.de
\addr Kiel University
\AND Johannes Brünger
\email johannes.bruenger@ibak.de
\addr IBAK GmbH
\AND Reinard Koch 
\email rk@informatik.uni-kiel.de
\addr Kiel University
}

\editor{}

\maketitle

\begin{abstract}

In the field of image classification, existing methods often struggle with biased or ambiguous data, a prevalent issue in real-world scenarios. Current strategies, including semi-supervised learning and class blending, offer partial solutions but lack a definitive resolution. Addressing this gap, our paper introduces a novel strategy for generating high-quality labels in challenging datasets. Central to our approach is a clearly designed flowchart, based on a broad literature review, which enables the creation of reliable labels. We validate our methodology through a rigorous real-world test case in the biomedical field, specifically in 
 deducing height reduction from vertebral imaging. Our empirical study, leveraging over 250,000 annotations, demonstrates the effectiveness of our strategies decisions compared to their alternatives.

\end{abstract}

\begin{keywords}
Data quality management, ambiguous data, annotation data application, data-centric AI, vertebral fracture
\end{keywords}

\section{Introduction}

Deep learning methods are at the leading methods for image classification, dependent on the availability of substantial high-quality data~\citep{are_we_done,relabelImagenet}.
While techniques like self- or semi-supervision can reduce the need for labeled data, high-quality labels remain a necessity, especially in domain-specific tasks~\citep{Liu2023React}.

A critical challenge lies in the reliability and consistency of human annotations, especially for ambiguous or complex classification tasks.
The consensus in current research suggests that single annotations are inadequate for ensuring label quality in such cases~\citep{Davani2022BeyondMajority,grossmann2022beyond,Basile2021ConsiderDisagree,Sharmanska2016AmbiguityHelps}.
To mitigate this, the adoption of soft labels, derived from averaging multiple annotations, has been proposed~\citep{schmarje2022benchmark}.
Soft labels can capture the inherent data uncertainty, which is different from the model uncertainty often assumed by uncertainty estimation methods.
However, obtaining multiple annotations per image is resource-intensive and impractical for large datasets or when expertise is scarce~\citep{imagenet}.
\citet{schmarje2023spa} introduces proposal-guided annotations, where a pre-trained network provides preliminary class estimates to guide annotators, thereby enhancing annotation efficiency and quality~\citep{desmond2021semi,foc,dc3}.
An illustrative example of this process is shown in \autoref{fig:ambiguous}.

In this paper, we present a comprehensive strategy for generating high-quality data in ambiguous classification scenarios.
High quality data means that we improve the label quality in comparison to one annotation only.
We validate our approach in vertebral fracture diagnosis, a field with significant challenges due to data ambiguity, which is ideal for testing our strategy's efficacy.

\picLarge{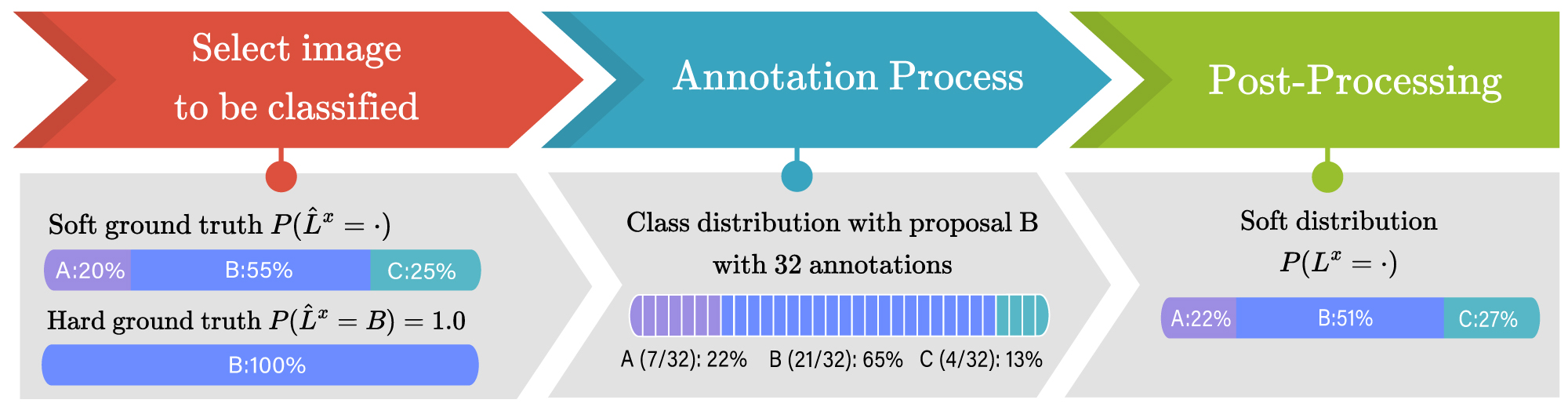}{Illustration of the concept of hard and soft labels and how they can be created from annotations -- 
The recommended process has three steps. 
In the first step, an image $x$ is selected for annotation. 
The unknown ground-truth distribution ($P(\hat{L}^x = \cdot)$) could either be soft or hard as shown by the examples in the lower half. 
During the annotation, multiple annotations are created either with or without proposal. 
A proposal means that one class is recommended during the annotation process. 
In the example, class B is proposed and 32 annotations are generated. 
The average across these annotations could already be used as an approximation of the soft-label $P(L^x = \cdot)$, however it might be biased towards the proposal since it is more likely to accept a proposal~\cite{schmarje2023spa}.
Our post-processing step enhances the approximated distribution ($P(L^x = \cdot)$)from the second step by reducing this bias. 
In the provided example, the probability of class B is reduced since it was most likely overestimated due to the used proposal of class B.}{1.0}

The diagnosis of vertebral fractures holds critical importance in medicine.
As noted by~\citet{haczynski2001vertebral}, vertebral fractures significantly increase mortality risks and recurrence rates.
Classifying these fractures, primarily based on vertebral height reduction, is challenging due to degenerative changes, leading to inconsistent annotations and impacting neural network training.
This scenario underscores the necessity of our approach, especially when domain experts are scarce.

Our primary contribution lies in synthesizing literature insights into a practical, step-by-step annotation guide for ambiguous real-world data.
While individual annotation techniques have been explored previously, our unified strategy addresses the entire annotation process comprehensively.
We aim to make this approach universally applicable to various image classification tasks, ensuring high-quality data.
Furthermore, we will release software guidelines to assist in creating high-quality annotations following our approach.

Our second key contribution is empirically validating our strategy on a vertebral fracture dataset~\citep{Loffler2020Verse2019,Sekuboyina2021Verse2020}.
By defining the guidelines prior to dataset testing, we provide an objective evaluation, underscoring the real-world applicability of our method.

By focusing on practical applications, this work seeks to promote a data-centric perspective in research and application, contributing to the discourse on data quality best practices.

\section{Practical Guidelines: How to Annotate Ambiguous Data?}
\label{sec:howannotate}

Before detailing our approach, it's essential to understand why these guidelines are crucial, especially in the context of ambiguous data.
Typical labeling guidelines~\citep{Sager2021SurveyLabeling,Chang2017Revolt} rely on definitive, hard-encoded labels and often overlook the disagreement among annotators about a given image's class.
Such discrepancies lead to ambiguous labels or, in broader terms, ambiguous data -- a prevalent issue in numerous real-world applications~\citep{Tarling2021UnderwaterUncertainty,Sambyal2021MedicalUncertainty,cancergrading,tailception,Jiang2021HarmfulContent,Zhang2023SoftSemantic}.
The probability distribution of a hard label for an image $x$ can be represented as $P(L^x = \cdot), P(\hat{L}^x = \cdot)  \in \{0,1\}^K$, where $K$ is the number of classes and $L^x, \hat{L}^x$ are random variables mapping an image $x$ to its class probability.
We distinguish between $L^x$ and $\hat{L}^x$, which represent the estimated and ground-truth probability distributions for image $x$, respectively.
The literature underlines that a single annotation per image is inadequate for capturing data ambiguity~\citep{Uma2021SurveyDisagreement,schmarje2022benchmark,schmarje2023spa,Davani2022BeyondMajority,grossmann2022beyond,Basile2021ConsiderDisagree,Sharmanska2016AmbiguityHelps}, necessitating multiple annotations for high-quality data.

\citet{Davani2022BeyondMajority} demonstrated that averaging multiple annotations to form a soft label, as proposed in \citep{schmarje2022benchmark,Hemming2018IDEA}, is superior to using a majority vote in ambiguous cases.
Moreover, label smoothing, a technique that tempers hard-coded labels with a constant factor, has been established as a method for enhancing network performance~\citep{Vaswani2017Attention,krothapalli2020adaptive,lukasik2020smoothing}.
A soft-label probability distribution for image $x$ can be expressed as $P(L^x = \cdot), P(\hat{L}^x = \cdot) \in [0,1]^K$.

It is crucial to recognize that ambiguous data is not merely an error, as suggested in~\citep{Park2022PseudoLabelingAmbiguous}, but a characteristic of the data itself.
Such data inherently possess uncertainty and should be treated accordingly.
As such, the soft ground truth distribution $P(\hat{L}^x = \cdot)$ is generally unknown and can only be approximated for validating predicted results.
This approach aligns with the growing emphasis on data-centric deep learning~\citep{Liu2021AutoDC,Jarrahi2022DataCentric,Whang2023DataCentric,grossmann2022beyond}, which prioritizes data quality over model architecture.

We present a comprehensive overview of our practical guidelines through a flowchart in \autoref{fig:flowchart2}.
The guidelines encompass five main steps: defining the specifics of what, who, and how the annotation process will be executed, the annotation process itself, and its post-processing.
The subsequent sections provide insights into the pipeline, with detailed discussions in \autoref{app:flow} and \autoref{subsec:details}.

\picLarge{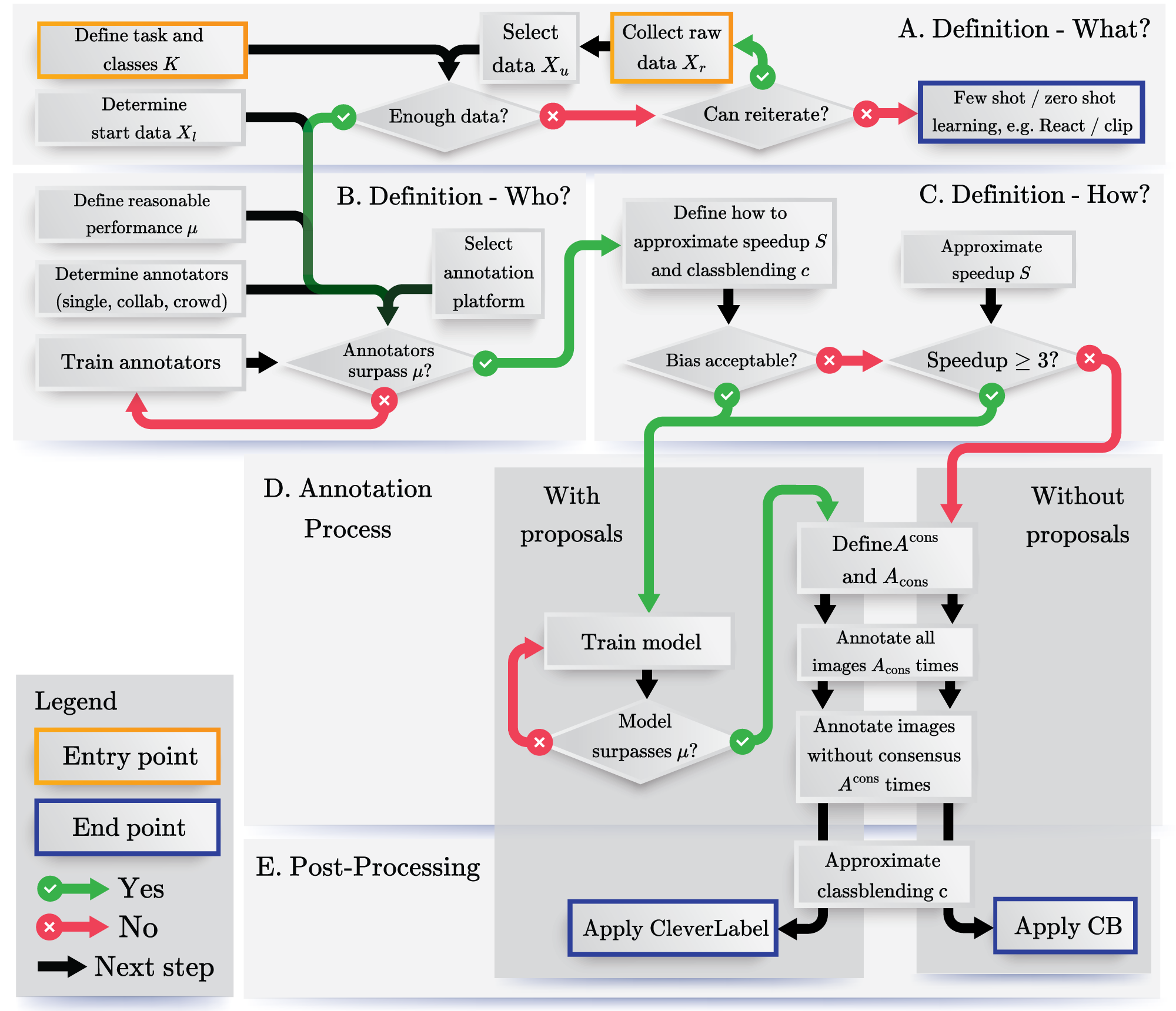}{Flowchart with guidelines on how to annotate ambiguous data, best viewed in color.}{0.99}

\begin{enumerate}
\item \textbf{Definition - What?} \
This stage outlines the classification task, specifying the classes ($K$) and gathering raw, unlabeled image data ($X_r$). A crucial element is selecting a representative subset ($X_u$) for annotation. The selection of $X_u$ emphasizes the importance of having an adequate sample size per class for effective training and evaluation or considers the use of self-supervised techniques for datasets without enough representative information. The definition of $X_l$ describes a smaller, precisely labeled dataset for evaluating annotators and training proposal generation networks.

\item \textbf{Definition - Who?} \\
This stage focuses on identifying suitable annotators for image labeling, ranging from individuals to crowdsourcing.
It is important to train the annotators according to the task's specific needs and setting a quality threshold for annotations ($\mu$), typically recommended between 60\% and 80\%. For classes with higher occurrence rates, a browsing environment for annotation is advised to improve process efficiency.

\item \textbf{Definition - How?} \\
In this stage, the decision to include proposals in the annotation process is discussed. Proposals can accelerate annotation but may introduce bias as discussed in \citep{schmarje2023spa}. Key considerations include the acceptability of bias, the extent of speedup ($S$)), and a class confusion matrix ($c$) for later processing. Proposals are recommended if the bias is manageable and the speedup significant (typically above a threshold of 3), otherwise, standard annotations are preferred.

\item \textbf{Annotation Process} \\
This phase involves conducting the annotations, with or without proposals. For ambiguous data, overclustering~\citep{schmarje2021datacentric,foc} is advised, and DC3~\citep{dc3} proposals are recommended for better performance. We propose to separate the annotations process into the number of annotations needed for early consensus ($A_\text{cons}$) and the total annotations for difficult cases($A^\text{cons}$). This separation balances ground-truth accuracy against the effort and cost of obtaining annotations.

\item \textbf{Post-Processing} \\
The concluding phase addresses the bias potentially introduced by proposals in the annotation process. CleverLabel~\citep{schmarje2023spa}, combining class blending and bias correction, is proposed for refining the approximated distribution if proposals were used. Even without proposals, blending the distribution with the class distribution ($c$) is recommended to enhance it.

\end{enumerate}

\section{Evaluation}
\label{sec:evaluation}

Our objective is to address the challenges of annotating real-world application tasks, such as classifying vertebral fractures in medical images.
To demonstrate the practicality of our proposed strategy, we apply it to the task using publicly available datasets Verse2019\citep{Loffler2020Verse2019} and Verse2020\citep{Sekuboyina2021Verse2020}.

\subsection{Applying the Strategy}
\label{subsec:apply}

\begin{figure}
\begin{minipage}[t]{0.54\linewidth}  
\vspace{0pt} 
    \centering
    \includegraphics[width=0.99\linewidth]{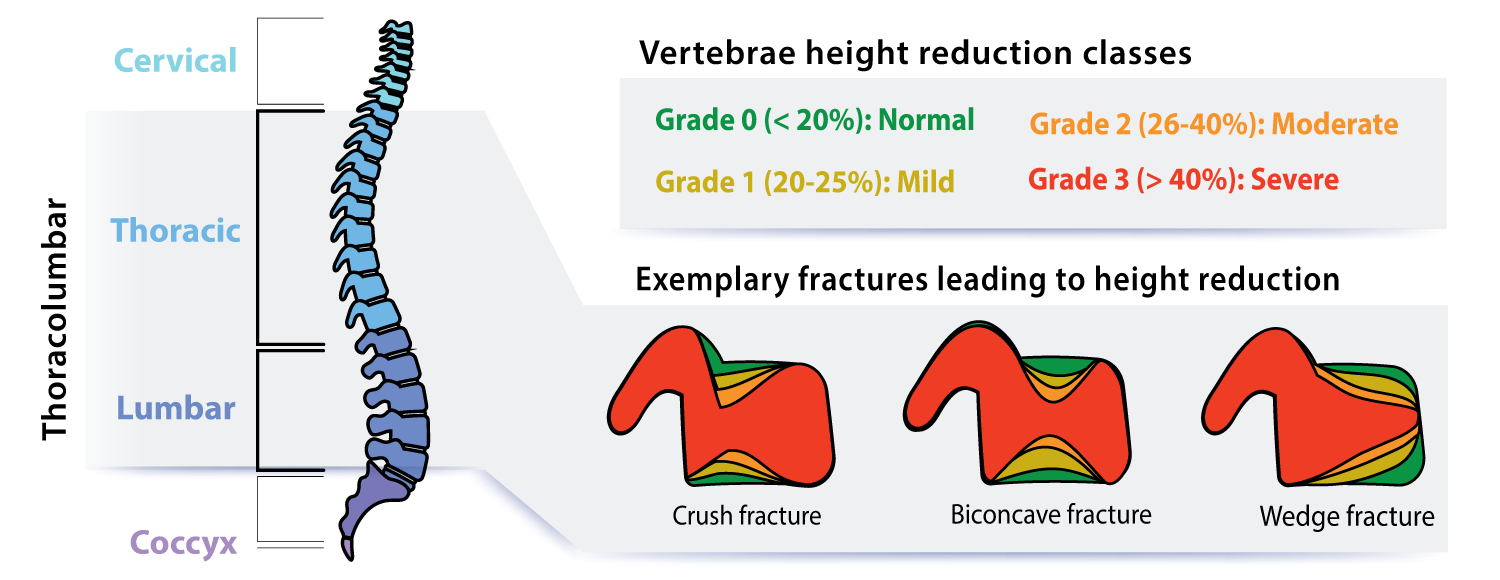}
    \caption{Illustration of a spine and definition of height reduction classes}
    \label{fig:vertebral}
\end{minipage}%
\hfill
\begin{minipage}[t]{0.45\linewidth}  
    \vspace{0pt} 
    \centering
    \resizebox{0.99\linewidth}{!}{%
    \begin{tabular}{l c  c    c c c c}
        \toprule
        Name             &   $\delta$ & $\hat{p_c}$ & \# Classes & Largest Class [\%] &  $S$ \\
        \midrule
        Verse (Ours) & 0.1143 & 0.6833 & 4 & 90.11 & 1.1636 \\
        CIFAR10-H & 0.0973 & 0.7766 & 10 & 10.16 &  2.4665 \\
        MiceBone & 0.3636 &  0.4775 & 3 & 70.48 & 1.4471 \\
        Plankton & 0.5784 & 0.7368 & 10 & 30.37 & 4.4319 \\
        Turkey & 0.2164 & 0.6863 & 3 & 75.95 & 1.4877 \\
        \bottomrule
    \end{tabular}
    }
    \caption{Comparison of dataset-specific variables with previously reported values \citep{schmarje2023spa}}
    \label{tbl:datasets} 
\end{minipage}
\end{figure}

The Verse datasets were chosen for their reproducibility and suitability in validating our workflow for classifying osteoporotic vertebral fractures. 
Following \citet{Sekuboyina2021Verse2020,Loffler2020Verse2019}, we focus on thoracolumbar vertebrae, the primary site for osteoporotic fractures, as illustrated in \autoref{fig:vertebral}.
Our classification approach is based on an adapted version of the Genant semi-quantitative score~\citep{Genant1996}, which categorizes fractures into four classes based on the degree of height reduction in vertebrae compared to their neighbors.
However, we exclude degenerative deformities, such as minor height reductions (up to 20\%), which are often indistinguishable in the given CT images due to lower resolution compared to standard radiographs.
As detailed in \autoref{tbl:datasets}, this results in a dataset of 3,761 individual vertebrae, with a notable class imbalance skewed towards class zero.

In contrast to the original authors of the Verse2019 dataset \citep{Loffler2020Verse2019}, we employ non-medical experts for annotation and utilize 2D projections from the 3D CT data.
The rationale and specifics of this approach are elaborated in the appendix \autoref{app:difference_previous}.
Further details regarding data preprocessing, annotators, annotation platform, approximated variables, and used hyperparameters are provided in \autoref{app:variables}.

\subsection{Analysis}
\label{subsec:results}

Our annotators completed over 100 iterations, half with proposals and half without.
However, due to the failure of one annotator to meet the acceptance threshold $\mu$, their annotations, including training iterations, were excluded, leaving us with 80 valid iterations or about 250,000 annotations.
An additional 11 iterations employed 3D data from the test dataset for comparative analysis.
The average annotation rate was calculated at 3,259.36 annotations per hour, indicating that approximately two hours are needed to complete the 3,761 annotations, accounting for technical overhead and breaks.
Comparison of dataset-specific variables with those in previous studies \citep{dc3} is provided in \autoref{tbl:datasets}.
 $\delta$ represents the data-specific offset when annotating with proposals as described in \citep{schmarje2023spa} and represents how strongly annotators are influenced in their decision by a proposal. 
 0 means no influence while 1 means following the proposal completely.
        $S$ denotes the speedup between annotating with and without proposals.
        $\hat{p_c}$ indicates the percentage of data with at least 95\% consistent annotations.
Notably, the dataset offset $\delta$ and near-consensus rate $\hat{p_c}$ are similar to prior reports.
We observed a larger dominant class and a smaller speedup $S$ in our dataset, as shown in \autoref{tbl:datasets}.
This finding is theoretically consistent with the notion that a proposal is less impactful when the majority of images already belong to a specific class, as is the case with the high class imbalance towards class 0 in the Verse dataset.
Our human annotator analysis indicated comparable quality to original test data, with increased uncertainties in cases of challenging or incorrect majority votes. Further details are provided in \autoref{app:human_analysis}.

\subsubsection*{Evaluation of Network Performance with Hard Labels}

Firstly, we evaluate the performance of our approach on the originally defined test data by \citep{Loffler2020Verse2019}.
Our best performing network for proposal generation achieved a macro F1 score of 0.57.
This network was a semi-supervised Mean-Teacher model \citep{mean-teacher} with overclustering and default hyperparameters from \citep{dc3} and was trained on the original annotations on the subset $X_l$ as defined in \autoref{app:difference_previous}.
A standard ResNet50 model \citep{resnet} with default hyperparameters (details in the supplementary) and cross-entropy loss function attained a score of 0.63 if it was trained not on our newly annotated data.
This outperformed the more complex proposal network and prior results by \citet{Wei2022Verse} for non-specialized networks.
Our data-centric approach enabled these improvements without model modifications, purely based on input data quality.
Future research could explore further enhancements by incorporating specialized loss functions and advanced backend models into our workflow.

\tblDeepLearningResults

\subsubsection*{Evaluation of Network Performance with Soft Labels}

In \autoref{tbl:dlResults}, we compare the network predictions on the improved distributions with blending and correction (($P(L^x = \cdot)$) against the approximated soft ground truth distribution ($P(\hat{L}^x = \cdot)$) using Kullback-Leibler divergence \citep{kullback}.
Comparing DC3 proposals with and without the proposed blending and correction steps in \citep{schmarje2023spa}, we find improved results using both methods.
Therefore, we confirm the efficacy of CleverLabel (a combination of these improvements) in the post-processing phase.
Additionally, we can validate that applying balanced blending to annotations without proposals enhances results.
Particularly, class blending based on the majority class proved effective for this task (Only Blends).
However, the benefit diminishes with more than five annotations, at which point it becomes less effective than our strategy's recommended method.
In applying our proposed annotation strategy (\autoref{subsec:apply}), we noted that bias introduction was not a major concern.
Yet, our results affirm that the strategy leads to optimal outcomes even in scenarios where bias is unacceptable.
Given a speedup of about 1.2, our flow chart would recommend to annotate without proposals, using only balanced blending, which emerged as the most effective method for multiple annotations.
Theoretically, with higher speedups, we could compare a larger number of annotations with proposals to fewer annotations without proposals.
For instance, at a decision boundary speedup of 3, five annotations with a proposal have a lower annotation cost than three without proposal (5 $\cdot$ 0.33 vs. 3) and yield a reduced KL score.
Hence, our strategy recommends the correct approach even in scenarios with higher speedups.

\subsection{Limitations and Future Work}\label{sec:limits}

The use of a single dataset for validating our strategy may appear as a limitation, and the reliance on existing literature could be seen as a weakness.
However, we contend that these aspects are, in fact, complementary.
Our strategy, derived from a synthesis of recent research, leverages verified methodologies to ensure the validity of the overall approach, rather than just its individual components.
Moreover, all potential decision paths in the graph were empirically tested.
Thus, we believe the validation using the Verse datasets is adequate, chosen for their real-world relevance and abundance of ambiguous data, necessitating complex annotations and yielding insightful results.
While the Verse datasets are smaller and less widely used than ImageNet~\citep{ILSVRC15} or CIFAR10~\citep{cifar}, their advantage lies in providing non-curated, class-distinct datasets.
Although curated datasets like ImageNet include ambiguous classes~\citep{are_we_done,Vasudevan2022DoughBagel,cifar10h}, they do not exhibit the level of noise and ambiguity found in real-world data as demonstrated in \citep{foc,schmarje2022benchmark}.
The requirement for multiple annotations under various setups for this study necessitated over 250,000 annotations for roughly 4,000 images, making such an extensive evaluation unfeasible for larger datasets at present.
While it is easier to annotate any classification task with our strategy, we still need many more annotations for evaluation, and thus evaluation is currently limited to smaller datasets.
Future research aims to apply and validate this approach on a larger scale, and we invite fellow researchers to join us in this endeavor.

\section{Conclusion}

This study introduces a comprehensive strategy for annotating and processing ambiguous data, with a specific focus on vertebral fracture diagnosis.
Our analysis of human annotator performance underscored the efficacy of proposal-guided annotation, which not only enhanced F1 scores but also streamlined the training process.
A neural network trained on the newly annotated data exhibited superior classification performance on the original test set, potentially halving the Kullback-Leibler (KL) divergence score compared to our new baseline.
Through thorough evaluation across all possible scenarios, we have substantiated the effectiveness of our strategy, rooted in literature, for this specific application.
By providing practical guidelines and demonstrating their successful application in a real-world context, this research significantly contributes to the creation of high-quality datasets and the advancement of data-centric deep learning.

\section{Reproducibility Statement}

To ensure reproducibility, we have documented and made accessible all essential resources of this study.
Our source code, data processing methods, model training, and evaluation procedures, along with a model card, are provided for transparency.
We utilized the publicly available Verse'19 \citep{Loffler2020Verse2019} and Verse'20 \citep{Sekuboyina2021Verse2020} datasets, concentrating on thoracolumbar vertebrae for classifying osteoporotic fractures.
Detailed preprocessing instructions are included to help with data reconstruction.

The human annotation process was executed on a web-based platform, employing non-medical expert annotators who received specific training for this task.
This project entailed over 250,000 annotations across approximately 4,000 images.
To minimize variability inherent in human annotation, we established comprehensive guidelines and criteria for a consistent approach.

For model training, we adopted a semi-supervised Mean-Teacher \citep{mean-teacher} model with overclustering, utilizing a ResNet50 architecture.
Key hyperparameters included a learning rate of 0.03 for the proposal network and 0.1 for the evaluation network, batch sizes of 64 and 128, and a weight decay of 0.0005.
The proposal network underwent training for approximately 600 epochs, while the evaluation network was trained for about 60 epochs.

\section*{Acknowledgments}

We acknowledge funding of L. Schmarje by the ARTEMIS project (Grant number 01EC1908E) funded by the Federal Ministry of Education and Research (BMBF, Germany).
We further acknowledge the funding of V. Grossmann by the Marispace-X project (grant number 68GX21002E), both funded by the Federal Ministry for Economic Affairs and Climate Action (BMWK, Germany).

\bibliography{lib}
\newpage

\appendix
\section{Full Explanation of Flowchart}
\label{app:flow}

\subsection{Definition - What?}
\label{subsec:what}

The first crucial step involves defining the classification task. Specifically, this entails identifying the $K$ classes relevant to the given problem and dataset. Additionally, gathering or generating raw, unlabeled data, denoted as $X_r$, is necessary based on the use case.

The next step involves deciding the number of images to label. We consider use cases with several thousand to tens of thousands of images, a common scenario in real-world applications~\citep{tailception,foc,volkmann_turkeys,benthic_uncertainty,deep_fish}. For datasets with hundreds of thousands or millions of images, self-supervised techniques like SimCLRv2 or OmniMAE~\citep{simclrv2,girdhar2022omnimae} may be applicable. However, even these cases require a smaller labeled dataset for performance evaluation, which can be labeled using our strategy. We denote $X_u \subseteq X_r$ as the subset containing samples to be labeled.

Our focus is on common supervised or semi-supervised deep learning approaches~\citep{fixmatch,mean-teacher,pseudolabel}. Therefore, it's imperative to ensure enough samples per class are available post-annotation. The term "enough samples" varies depending on the use case. Based on previous results~\citep{schmarje2022benchmark}, each class should represent at least 1\% of the complete data, with several dozen images per class for robust evaluation. In cases of limited images, additional considerations for few or zero-shot learning approaches are necessary~\citep{survey-few,survey-zero,zeroshot}, although this extends beyond our paper's scope.

For each image $x$ in a small labeled dataset $x \in X_l \subseteq X_u$, at least one hard-encoded ground truth label is necessary. This data is crucial for annotator evaluation (see \autoref{subsec:who}) and potentially for training a proposal generation network (see \autoref{subsec:how}). An initial dataset size of about 20\% of the total dataset size was beneficial in \citep{schmarje2023spa}, though the exact size may vary depending on the specific use-case.

\subsection{Definition - Who?}
\label{subsec:who}

The choice of annotators is a pivotal aspect of the annotation process. Various approaches include using a single user, collaborating with several workers, or employing crowd-sourcing methods~\citep{Sager2021SurveyLabeling}. The annotators' background dictates the extent of training or selection necessary for the task~\citep{effifient_annotation}.

For evaluating annotation quality, two elements are essential alongside the small labeled dataset $X_l$: a performance threshold $\mu$ and an appropriate annotation platform. The threshold $\mu$ should align with the evaluation metrics used, such as macro accuracy or F1-score, which often show high correlation~\citep{temporal-ensembling,He2021MaskedAutoencoder,mixmatch,schmarje2022benchmark,temerateFish,divide-mix}. \citep{dc3} suggests that a threshold between 60\% and 80\% is typically adequate if annotations are aggregated. However, the label quality in $X_l$ influences the maximum achievable scores; for instance, scores higher than 80\% might be unattainable due to intra- or inter-observer variability~\citep{schmarje2019}.

Based on the task definition (see \autoref{subsec:what}), a browsing environment is recommended for tasks with a few classes having many occurrences, allowing for side-by-side image comparison~\citep{Sager2021SurveyLabeling}. Alternative platforms like \citep{morphocluster} are also viable. Although these platforms may require setup overhead, they can significantly expedite the annotation process. Annotators should be trained to exceed the performance threshold $\mu$ on the labeled data $X_l$ using the chosen platform.
Another concern might be the annotator calibration, which is the alignment of annotations between the annotators. 
We actually do not require such alignment, because we measure these differences between the annotators. 
However, if an easy rule allows better alignment of annotators and yields less ambiguous results this direction can be pursued.

\subsection{Definition - How?}
\label{subsec:how}

A critical decision in this stage is whether to incorporate proposals in the annotation process. While proposals can significantly reduce annotation costs, they may introduce a bias~\citep{jachimowicz2019default,schmarje2023spa}. This bias results from annotators' tendency to favor the proposed class, skewing the label distribution $P(L^x = \cdot)$ towards it. For making this decision, several parameters, such as the class confusion matrix $c$ and the expected annotation speedup $S$, need estimation or a defined approach for later approximation (see \autoref{sup:workload} and \autoref{sup:conf} for techniques).

The first question to address is whether introducing a bias is acceptable. In cases where bias is tolerable, proposals should be used. If not, the expected speedup $S$ becomes the determining factor. Proposals are more advantageous with a higher speedup, as a higher speedup allows for easier reversal of the introduced bias with more annotations. \citep{schmarje2023spa} identifies a trade-off point for speedup around 3; below this, it's advisable to avoid proposals due to the difficulty in reversing the bias within a reasonable budget. The class confusion matrix $c$ is utilized in post-processing (see \autoref{subsec:post}) and not needed until then.

\subsection{Annotation Process}

This step focuses on creating the annotations, with various preparations depending on whether proposals are used. Literature indicates that proposals, particularly DC3 proposals, enhance annotation consistency, especially with ambiguous data~\citep{papadopoulos2021scaling,desmond2021semi,foc}. We recommend DC3 proposals~\citep{dc3}, which perform well with real-world ambiguous data and are compatible with many semi-supervised methods~\citep{fixmatch,divide-mix,mean-teacher}. Normal network predictions can also be used but might reduce overall performance. The labeled dataset $X_l$ serves as training data for a neural network to generate these proposals, which should be trained until reaching the same quality threshold $\mu$ as the annotators.

Two important variables to define are the number of annotations required for consensus ($A_\text{cons}$) and the total number of annotations ($A^\text{cons}$). These variables balance between the best approximation of the ground-truth distribution $P(\hat{L}^x = \cdot)$ and the effort and cost of annotations. In \autoref{sup:workload} and \autoref{sup:conf}, we discuss plausible numbers of annotations per hour and potential confidence guarantees for different combinations of $A_\text{cons}$ and $A^\text{cons}$.
The annotations can either be created by different individuals or by showing the same images multiple times to one or multiple annotators.
For the latter case, it is important to think about a cooldown period, the time between showing the same image again. 
This period can be quite small, down to a few hours, as long as there is a high probability that the annotator will not remember a case between annotations.

\subsection{Post-Processing}
\label{subsec:post}

The use of proposals in annotation can introduce biases due to the default effect \citep{jachimowicz2019default}. \citet{schmarje2023spa} propose CleverLabel, a combination of class blending and bias correction, to improve the approximated distribution if proposals were used. If no proposals were employed, enhancing the distribution $P(L^x = \cdot)$ by blending with the class distribution $c$ is still recommended.

\section{Implementation and approximation details}
\label{subsec:details}

This subsection provides more in-depth descriptions for some of the previous parts. 
They are separated to allow a concentration on the most important information without diluting it with necessary but minor considerations. 
We start this section with a description of additional minor consideration which apply to the complete annotation strategy (see \autoref{subsubsec:considerations}).
Additionally, we will discuss how the expected workload for the human annotators can be calculated (see \autoref{sup:workload}) and what confidence with regard to the soft label can be expected given a certain number of annotations (see \autoref{sup:conf}).
Lastly, we will reiterate the most important information about CleverLabel from \citep{schmarje2023spa}.

 \subsection{Additional minor considerations for the Annotation Strategy}
 \label{subsubsec:considerations}

\begin{itemize}
    \item  All decisions should typically be supported by domain experts or the end users which will use the labeled data in the end.
    \item Constraints like collaboration modes and support for proposals should be considered.
For example, if crowd-sourcing should be used for annotation, this restricts the appropriate platforms.
Due to the fact the decision if proposals will be used is made later in \autoref{subsec:how}, it is advisable to ensure that these future options are supported to avoid unnecessary backtracking.
\item If the training of the annotators is not successful it might be necessary to step back and rethink previous decision to allow a successful training. 
For example, special training with experts could be held, more detailed description of the task could be given and general feedback of annotators could be considered.
\item  Proposals $\rho_x$ with a low ground truth probability ($P(\hat{L}^x = \rho_x) = 0$) should be avoided and \citep{dc3} indicates only up to 10\% of all proposals should be of such poor quality to allow successful training.
The distribution $P(\hat{L}^x = \cdot)$ might not be known at this point but if available this should be checked in addition to the threshold $\mu$.
\item  The variables $c, \delta$ and $S$ may be approximated in various ways.
One possible approach is defined in \citep{schmarje2023spa} by annotating  about 100 images at least 10 times for the estimation of $c$, $\delta$ and $P(\hat{L}^x = \cdot)$.
The speedup $S$ can for example be measured on a small subset with and without proposals during the training of the annotators.
However, if the variables are not required during the decision process described above, they could also be approximated on the real annotation data or ignored entirely.
For example, $c$ can easily be approximated on the complete dataset if no proposals are used during the annotation process and an approximation of $S$ is not required if a small introduction of a bias is acceptable.
The variables could also be approximated with an educated guess either based on previous knowledge or experiments.
\item The question is how $A_\text{cons}$ and $A^\text{cons}$ should be selected.
We concentrate on two major aspects, what is a feasible workload for the annotators and what guarantees can be given. See for more details the approximations in \autoref{sup:workload} and \autoref{sup:conf}.
\item The class blending (CB or CleverLabel) in the post-processing step may not be necessary if the number of annotations is high (e.g. above 50), since the distribution is already approximated quite accurately.
\end{itemize}

\subsection{Approximation of workload}
\label{sup:workload}

The expected workload can be approximated as the expected number of annotations per hour done by one annotator as $\frac{a}{h}$.
Depending on the expected portion of complete consensus of the dataset $p_c$ and the total dataset size $N = |X_u|$, we can calculate the expected annotation time in hours with $A_N = N \cdot (p_c \cdot A_\text{cons} + (1-p_c) \cdot A^\text{cons}) / \frac{a}{h}$.
In  \autoref{fig:annotation-time} the required annotation time $A_N$ in days for $N = 10,000, A_\text{cons} = 10, A^\text{cons} = 50$ with varying $p_c$ is given as well as values reported in \citep{dc3}.
The reported values for $\frac{a}{h}$ are between 1,000 and 4,500 without proposals or 3,000 and 10,000 with proposals.
Thus, even in the worst-case the expected workdays with proposals are less than 10 days, or two weeks, which should be feasible in most cases especially when multiple annotators are available.
These values are examples and have of course to be adopted to the specific use-case.

\FloatBarrier

\pic{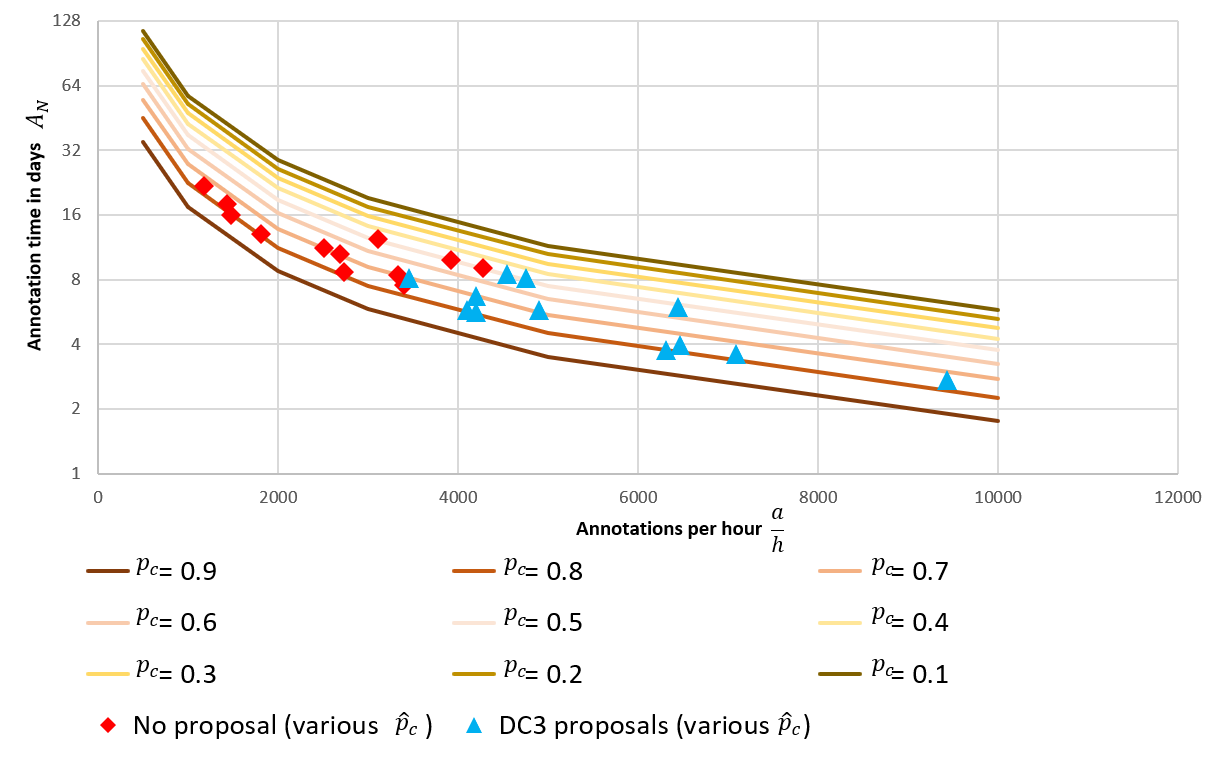}{Annotation time in days based on expected consensus ratio $p_c$ and annotations per hour $\frac{a}{h}$, reported values from \citep{dc3} and with 95\% agreeing votes threshold $\hat{p_c}$
}{0.9}

\subsection{Approximation of confidences}
\label{sup:conf}

The confidences of the calculated label distributions should be considered when $A^\text{cons}$ and $A_\text{cons}$ are defined.
Considering the binary case of the annotation task for class $k$ for each image $x$, i.e. whether class $k$ is annotated for $x$ or not, and having a sufficient number of annotations $A$, the 95\%-confidence interval can be calculated using the Wald confidence interval \citep{Lyman1993Statistic}.
The interval is given by
	\begin{equation}
	\label{eq:confidence_interval}
	\begin{split}
	P(\hat{L}^x = k) & \pm  Z_{95\%} \cdot \sqrt{\frac{p_k(1-p_k)}{A}}
	\end{split}
	\end{equation}
with 95\%-confidence value, which is 1.96, and $p_k$ is the unknown true GT probability for class $k$.
The confidence interval is the largest for $p = 0.5$.
The formula can be rearranged to calculate the required annotations $A$ if a 95\% confidence interval around the approximation $P(\hat{L}^x = k)$ of width $W$ and margin $W/2$ is to be guaranteed.
\begin{equation}
	\label{eq:confidence_number}
	\begin{split}
 \frac{W}{2} & = Z_{95\%} \cdot  \sqrt{\frac{p_k(1-p_k)}{A}} \\
  \frac{W^2}{4} & = Z_{95\%}^2  \cdot  \frac{p_k(1-p_k)}{A} \\
	A &=  \frac{4 \cdot Z_{95\%}^2 \cdot  p_k (1-p_k)}{W^2} \\
	\end{split}
	\end{equation}
\autoref{fig:confidences} visualizes the required annotations $A$ for various $p$ and $W$.
These calculations can guarantee with 95\% confidence that the confidence interval has a width of $W \approx 1$ for $A=3$,  $W \approx 0.62$ for $A=10$ and $W \approx 0.28$ for $A=50$.
However, a high number of $A$ is indirectly assumed because the Wald confidence interval approximates the original binomial distribution with a normal distribution.
This approximation is especially inaccurate for small numbers of $A$ and $p$ near 0 or 1, as pointed out by~\citet{Lyman1993Statistic}.
They suggest to approximate the 95\% confidence intervals with $((0.25)^{\frac{1}{A}}, 1)$ when $p$ is close to one.
This would give  a lower bound of the interval of $0.63$ for $A=3$, $0.87$ for $A=10$ and $0.97$ for $A=50$.
All these calculations assume an idealized version of the annotators, which must be taken into account in their interpretation.
For this reason, it will not be investigated further how the confidence intervals can be approximated more precisely, but conclude that a consensus decision is uncertain for $A=3$ annotations, but reasonable for ten or more annotations.
Similarly, approximations of $p$ with a confidence interval width less than 0.3 are only reasonable for about 50 annotations.

\pic{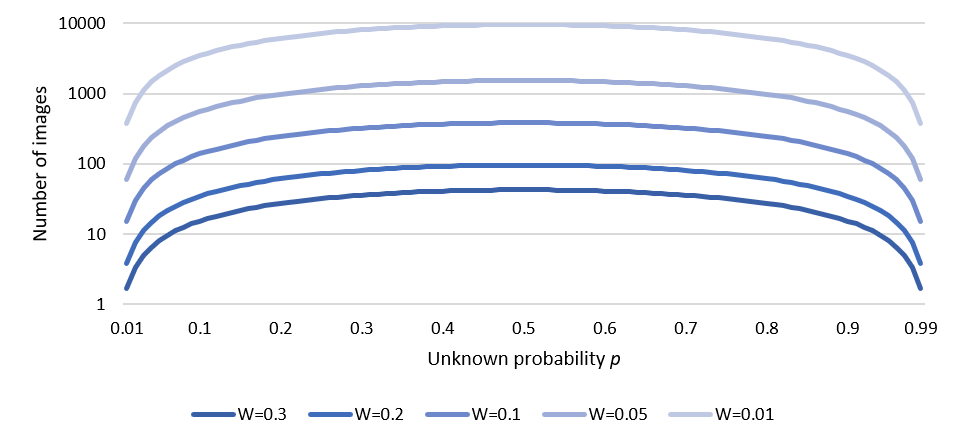}{Number of images for confidence width $W$, logarithmic scale}{0.9}

\subsection{Details on using CleverLabel and CB}

This section provides an overview about the methodology proposed in \citep{schmarje2023spa}. 
These elements are reiterated here to keep this paper self-contained.
Schmarje et al. presented in \citep{schmarje2023spa} two different approaches to improve the resulting distribution when using proposals.
The approaches were called class blending (CB) and bias correction (BC) and formed together the method CleverLabel.
Class blending uses a constant transition matrix $c$ between classes and blends this fixed class distribution with the calculated distribution.
The constant transition matrix is motivated by the fact that for example a car and truck are more easily confused than a car and a dog.
The bias correction used the theoretical analysis of the introduced bias when using proposals to invert this process. 
A main variable for this calculation is the dataset specific offset $\delta$.
This offset describes how likely it is that a human will accept the proposal even if the proposed class would normally not been considered. 
A offset of zero means that the proposal has no influence while a value of one means that the proposal is always accepted regardless of the content.
For details, please see the original publication~\citep{schmarje2023spa}.

\section{Differences to previous work}
\label{app:difference_previous}

\picFour{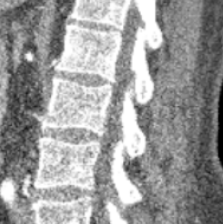}{CT slice}{CT projection}{Mask projection}{Blended}{Illustration of different image modalities}

Unlike to the original author of the Verse2019~\citep{Loffler2020Verse2019} dataset, we will not use medical experts for the annotation and will use 2D projections of the original 3D CT data.

Our decision to use non-medical experts has two main reasons.
First, we want to show that non-experts can successfully annotate data with the help of proposals.
Second, we plan to generate tens of thousands of annotations, which would not be feasible with medical experts, who are in high demand.
While we will not use medical experts for the majority of the annotations, we have worked closely with the radiologist at University Medical Center Schleswig-Holstein. 
We discussed our design choices with them, particularly regarding the definition of the task and the representation of the vertebrae used.
As mentioned above, we trained the hired workers on this specific task and ensured that they achieved a macro accuracy and F1 score of at least $\mu = 0.6$ on the complete dataset with and without proposals for at least two scores before they could annotate the final scores.

We do not use the original 3D CT data (see \autoref{fig:vps-vert-0}) for the annotation but rather a 2D projection of the central slices for each vertebra (see \autoref{fig:vps-vert-1}) and its segmentation (see \autoref{fig:vps-vert-2}).
Additionally, we show the vertebrae above and below the target vertebrae to allow an evaluation of the height reduction.
A blended example is shown in \autoref{fig:vps-vert-3}.
There are three major benefits/reasons of using 2D projections over 3D CT data.
First, the Genant scores were defined on 2D X-Ray and 2D projections of CT data are visually very similar to X-Ray and thus are better comparable.
 Second, we want to use the reported methods from the literature for the proposal generation~\citep{dc3}, which were only tested on 2D data.
 While it should be theoretically possible to extend the used networks to 3D, such major structural changes could lead to an inferior result which would dilute the results. But such experiments may be future work.
Lastly, when it comes to diagnostics, radiologists mainly look at one central slice which we include and the major 3D structures can be detected on this projection.

\subsection*{Annotators and annotation platform}

We hired 5 workers to annotate vertebrae in a web-based environment developed for this evaluation. 
Public annotation platforms do not support the combination of clusters and classes that we want to evaluate in this paper. 

\autoref{fig:website} displays the used annotation website.
It is important to note, that all images are displayed to the user in a grid.
In the case without proposal (see \autoref{fig:website-0}), the shown classes are mixed and all have to be annotated manually by the user.
In the second case with proposal (see \autoref{fig:website-1}), most of the images are of the same class and can be annotated simultaneously, while only the errors need to be corrected.
As $X_l$ we used the partially available labels from \citep{Loffler2020Verse2019}.

The software is a custom implementation to handle various inputs like classification and clusters.
We plan to release a version in the future and please contact us if you are interested in a preview.

\picTwoVertical{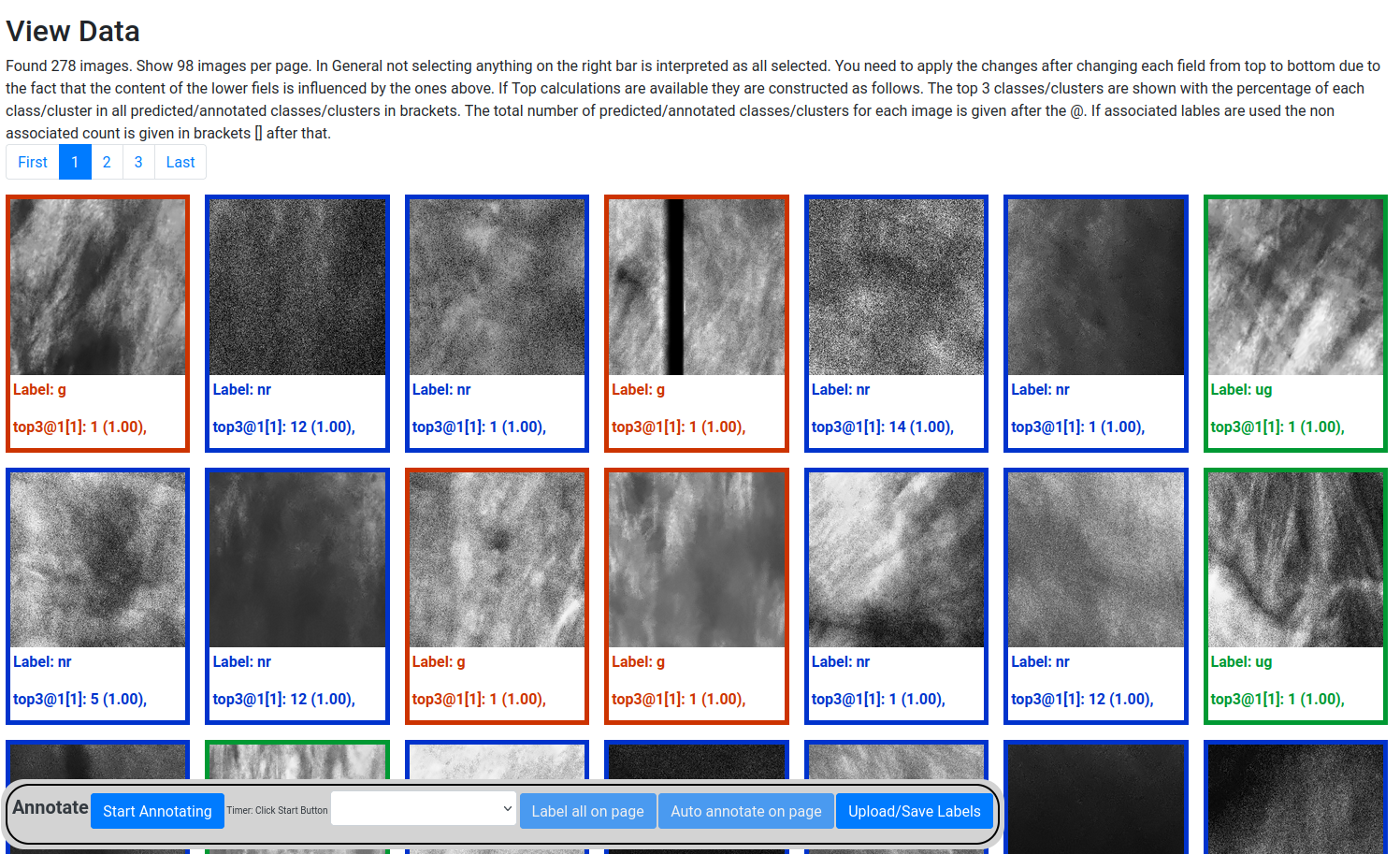}{No Proposal}{With DC3 proposal cluster}{Screenshots of the annotation website on the MiceBone dataset from \citep{schmarje2022benchmark}, The sidebar for data selection is hidden for data protection reasons. 
}

\section{Used parameters and their calculation}

\label{app:variables}
\subsection{Approximation of desired variables}

In a medical consensus process in an expert panel, the proposals of different people are unified, but influences between people due to social dynamics cannot be excluded.
Therefore, we argue that a bias that may be introduced due to human approved annotations is comparable to a consensus process and therefore acceptable.
This means that we will use proposals during the annotation and $c$ and $S$ only need to  be approximated after the annotation process based on the defined strategy.
To allow a validation of our decisions, we will also raise annotations without proposals and thus we can approximate the class distributions $c$ and speedup $S$ based on these annotations.
Additional approaches to approximate $S$ as well as the approximation of a data-specific offset $\delta$ (see \citep{schmarje2023spa}) are discussed in the supplement.

Based on the training annotations, we estimated the $\frac{a}{h}$ to be about 2,000 and 3,000 without and with proposals, respectively.
These results are similar values previously reported in \citep{dc3}.
For $A_\text{cons} = 10$, $A^\text{cons} = 50$ and estimated $p_c = 0.5$, this would result in about 250,000 annotations with and without proposals.
Furthermore, based on the estimated $\frac{a}{h}$ of about 2,500, this resulted in a workload of about 20 hours per annotator which was within our given budget.
So we chose the number of annotations at the recommended values of ${A_\text{cons}=10,~A^\text{cons}=50}$.
Additionally, we planned to annotate the 3D data to evaluate the quality difference between 2D and 3D. 
We limited these annotations to the test set of the original Verse2019 dataset~\citep{Loffler2020Verse2019}.

We used annotations created during the training of the annotators to estimate $\delta$.
We randomly picked 20 examples without consensus from the annotations without proposals. 
Based on these samples, the proposal per image $p_x$, the probabilities $P(L_x = \rho_x)$ with and without proposal and the formula for simulated proposal acceptance of \citep{schmarje2023spa}, we calculated $\delta \approx 0.11$.

\subsection{Used hyperparameters}

The used hyperparameters are defined in \autoref{tbl:paras}. For more detail see the respective source code \footnote{ \url{https://github.com/Emprime/dc3}, \url{https://github.com/Emprime/dcic}
}.

\begin{table}
\centering

	\caption{Used hyperparameters for the proposal and evaluation network, $\dagger$ approximated based on number of image iterations, epochs not counted
		    }
		 \label{tbl:paras} 
\resizebox{\linewidth}{!}{%
		\begin{tabular}{l c c }
			\toprule

Parameter             &  Proposal network & Evaluation network \\
\midrule
Architecture &  DC3\citep{dc3} + Mean-Teacher\citep{mean-teacher}  & ResNet50\citep{resnet}\\
Learning rate & 0.03 &  0.1 \\
Epochs & 600$\dagger$ & 60 \\ 
Optimizer & SGD & SGD \\
Batch size & 64 & 128  \\
Weight Decay & 0.0005 & 0.0005 \\
Pretrained Weights & N/A & ImageNet \\
Source code & \citep{dc3} & \citep{schmarje2022benchmark}\\

			\bottomrule
		\end{tabular}

		   }
\end{table}

\section{Additional results of analysis}

\subsection{Learning effect}

Regarding the learning effect of the annotators, the training iterations showed an increase in F1 scores over time.
The improvement in F1 scores across all annotators between the training and the last three iterations was 6.79 without proposals and 9.19 with proposals. 
When comparing the benefit of DC3 proposals versus no proposals, the average improvement in F1 scores was 3.27 for the first three iterations and 0.41 for the last three iterations.
The proposal still has a benefit in the end but the annotators can more successfully annotate event without them.
Moreover, the introduction of proposal-based annotations resulted in a reduction in training time which also diminished over time.
The first three iterations saw a reduction of 13.49 minutes, while the last three iterations experienced a decrease of 4.55 minutes.
Overall, we can conclude that the proposals helped to improve the quality and reduce the annotation time.
However, this effect was diminished over time while the performance with and without proposals increases.
Thus, we see a learning effect of the annotators.

\subsection{Evaluation of human performance}
\label{app:human_analysis}
\tblhumanResults

A comparison of the annotator and network performance on the Verse2019~\citep{Loffler2020Verse2019} test set is given in \autoref{tbl:originalResults}.
When comparing the macro F1 scores in 2D between annotations performed with and without proposals and those with proposal-based guidance, the latter achieved a significantly higher F1 score of about 2\%.
This means that the employment of proposal-based guidance resulted in improved annotations by the annotators.
The average macro F1 score for 3D annotations shows no significant difference compared to the 2D annotations, meaning that the modality difference was of no to little importance to the annotators as expected in \autoref{subsec:apply}.

The marginal class distribution on all vertebrae from the test set between the ground truth and the majority vote for 2D annotations was quite similar, with percentages of 80.33, 10.38, 6.01, and 3.28 versus 80.33, 10.93, 5.74, and 3.01, respectively.
In addition, only 17\% of the majority votes deviated from the ground truth. 
Comparing the average uncertainty based on $P(L^x = \cdot)$ for concurring and dissenting majority votes with the ground truth, we find a higher annotator uncertainty of 24.20\% for dissenting scores than for concurring scores at 7.78\%.
These results indicate that the annotators produced annotations of similar quality to the original test data, with higher uncertainties for potentially difficult or incorrect majority votes.

\section{Dataset Card}

\begin{table}[h]
\footnotesize
\centering
\begin{tabular}{l p{10cm}}
\hline
\textbf{Attribute} & \textbf{Details} \\
\hline
Dataset Name & VerSe 2D Projections with Soft Labels\\
\hline

Dataset URL & 
\url{https://zenodo.org/records/8115942} 
\newline \textbf{Original datasets:} \newline
\vspace{-5mm}\begin{itemize} \item \textbf{VerSe'19:} \url{https://osf.io/nqjyw/} \item \textbf{VerSe'20:} \url{https://osf.io/t98fz/} \end{itemize}\vspace{-5mm} \\
\hline
Description & The dataset contains 2D projection of the central slices for vertebra of the VerSe (Large Scale Vertebrae Segmentation Challenge) datasets. These are collections of multi-detector CT scans for vertebral labelling and segmentation, significant for its size, diversity, and detailed annotations. The data, collected in conjunction with MICCAI 2019 and 2020, is anonymized and includes a variety of normal and pathological cases. \\
\hline
Data Collection & \textbf{Domain:} Medical Imaging, 2D Projections of Spine Multi-detector Computed Tomography (MDCT) \newline 
\textbf{Annotation Details:} Soft-labels by evaluating proposal-based human non-expert annotations \newline \textbf{VerSe'19:} 160 CT scans from 141 patients, 1725 annotated vertebrae \newline \textbf{VerSe'20:} 374 CT scans from 335 patients, 4505 annotations \\
\hline
Data Preprocessing & Scans and segementation masks have been anonymized and converted to NIFTI format (https://nifti.nimh.nih.gov/) in the VerSe dataset. The resolution of the CT images was adjusted to limit computational demands for deep learning algorithms. Coordinates of vertebral body centroids per vertebral level are stored in JSON format. \\
\hline
Annotations & The proposed multi-step algorithmic framework has been applied to generate annotations with or without proposals by trained non-experts, including post-processing with CleverLabel and class blending.\\
\hline
Usage & Development and evaluation of algorithms for vertebral labelling and segmentation in spine CT scans. VerSe‘19 and VerSe‘20 were used in challenges to evaluate accuracy, precision, and the effectiveness of segmentation and fracture detection as part of MICCAI 2019 and 2020.\\
\hline
Known Limitations & While including multiple sites, the datasets might still not represent all possible variations in spine pathology and patient demographics.\newline \textbf{VerSe'19:} Exclusion of patients younger than 30 years and certain spinal conditions like bone metastasis\\
\hline
Ethical Considerations & Both VerSe studies were approved by the local institutional review board, which conducted a thorough review of the research methodology, ensuring adherence to ethical guidelines. All patient data included in this study were anonymized to protect the privacy and confidentiality of the individuals.  \\
\hline
Acknowledgements & VerSe has been supported by the European Research Council under Horizon2020 (GA637164–iBack–ERC–2014–STG) and NVIDIA. \\
\hline
\end{tabular}
\caption{Overview of the VerSe 2D Projections with Soft Labels}
\label{table:verse_dataset}
\end{table}

\end{document}